\documentclass[a4paper, 10pt]{article}
\usepackage{amsmath,amsfonts,amssymb}
\usepackage{a4}
\usepackage{float}
\usepackage{amsthm}
\usepackage{palatino}
\usepackage{fancyhdr}
\usepackage{latexsym}
\usepackage{graphicx}
\usepackage[latin1]{inputenc}
\usepackage{hyperref}

\providecommand{\keywords}[1]{\textbf{Keywords:} #1}
\begin{document}
\begin{center}
\Large
\textbf{Arabic Handwritten Text Line Dataset}
\vspace{0.3cm}
\normalsize
\sffamily
\\
\underline{H. BOUCHAL}    ,   A. BELAID\

\vspace{0.3cm}
\footnotesize
LIMED Laboratory, Faculty of Exact Sciences, University of Bejaia,
06000 Bejaia, Algeria, hakim.bouchal@univ-bejaia.dz ,

Data Science \& Applications Research
Unit - CERIST, LIMED Laboratory, 06000, Bejaia, Algeria,
abelaid@cerist.dz 

\vspace{0.5cm}
\small

\rmfamily
\end{center}

\vspace{0.5cm}
\normalsize

\textbf{Abstract}: Segmentation of Arabic manuscripts into lines of text and words is an important step to make recognition systems more efficient and accurate. The problem of segmentation into text lines is solved since there are carefully annotated dataset dedicated to this task. However, To the best of our knowledge, there are no dataset  annotating the word position of Arabic texts. In this paper, we present a new dataset specifically designed for historical Arabic script in which we annotate position in word level.  \\


\section{Introduction} 

\hspace{0.5cm}The recent advances in the area of computer vision are mainly based on the use of deep learning models \cite{article123}. The proper functioning of a model requires a large amount of data and the creation of a data set is a difficult and time consuming task. Indeed, in supervised learning, labeling is essential because the models must understand the input features to process them and return an accurate result in output.

In recent years, historical documents manuscripts have received a great amount of interest. Indeed, many libraries and museum archives are interested in the automatic treatment and preservation of a large number of historical documents with the help of imaging systems. For any document the segmentation in words is a crucial step for any recognition system, this task faces several challenges mainly the overlap between words, words that touch each other in the same line or two adjacent lines, in the literature several studies have been performed on word segmentation based on statistical measures \cite{article2,article3,article4} by classifying the spaces between connected components as a space between two words or a space within words, but these methods are not efficient for text lines with overlapping words as there is no distance between the words.

Arabic script is one of the most challenging to recognition scripts in the field of optical character recognition \cite{article5}, we propose a new dataset built from the pages documents of the RASM2019 dataset \cite{article6}, which is a collection of 120 original pages of historical scientific manuscripts transcribed into Arabic from Arab countries and other countries with Arab or Muslim communities. It contains pages that are distinguished by the presence of margins, diagrams, tables, and images in addition to text line areas that show fascinating variations in style and writing, The diversity of these manuscripts provides unique and exciting challenges.

Our goal is to propose a manuscript document dataset containing word position coordinates. We have started to exploit high-resolution document images from the RASM2019 dataset that only contain the position of text lines, so we believe that the addition of word position can motivate researchers to propose more solutions for segmentation and recognition tasks.

\section{Dataset overview }
\hspace{0.5cm} Compared to the Latin language few dataset have been created for Arabic manuscripts which leaves the Arabic language far behind the document image analysis systems for different tasks, such as layout analysis, baseline detection, handwriting recognition, author identification, and many others. We list some databases for historical Arabic documents publicly available below.

\subsection{IBN SINA}
\hspace{0.5cm} IBN SINA dataset \cite{IBN SINA} obtained from 51 folios and corresponds to 20722 connected components of the historical Arabic handwriting. Each connected component can be a character, a sub word or a word. These components are extracted after a prepossessing step to restore and enhance the handwritten page.

\subsection{VML-HD}
\hspace{0.5cm} the VML-HD dataset \cite{VMLHD} collected 680 pages from five different books by different writers in the years 1088-1451. and completely annotated at sub word level with the sequence of characters. This dataset can be used for word-spotting and recognition. 

\subsection{MHDID}
\hspace{0.5cm} Multi-distortion Historical Document Image Database (MHDID) \cite{MHDID} includes 335 images of historical documents collected from 130 books published between the 1st and 14th centuries. Is used to evaluate the quality of degraded documents and classify degradations into four categories, paper translucency, staining, readers' annotations and worn holes.

\subsection{BADAM}
\hspace{0.5cm} The BADAM dataset \cite{BADAM} consists of 400 images of scanned pages of annotated documents from four digital collections of Arabic and Persian manuscripts from various fields and periods. The lines of text are annotated with a single baseline extended over the majority of the text online.

\section{Description and properties of the KALIMA-Dataset}

\hspace{0.5cm}The KALIMA-Dataset contains text line images extracted using polygonal coordinate annotations enclosing them of the original database \cite{article6}, along with its field transcription. Followed by a step of manually delimiting the position of each word in a text line by a bounding rectangle and associating to each word the ground truth of its transcription.  \textbf{Figure 1} shows an example of delimiting words in a text line by delimiting boxes.

\begin{figure}[h]
    \centering
    \includegraphics [width=\linewidth] {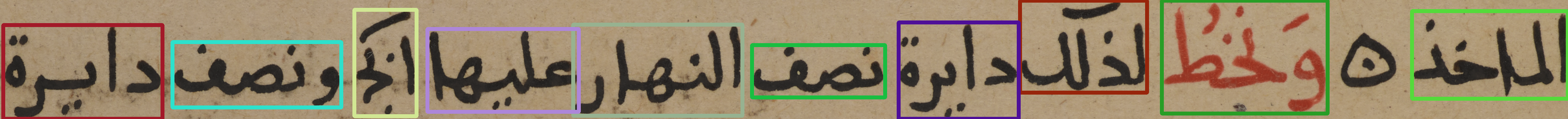}
    \caption{Example of delimiting words in a line of text with delimiting boxes.}
    \label{fig:1}
\end{figure}

\vspace{0.5cm}
The KALIMA dataset contains 497 images of text lines with 5062 bounding box annotations, of which 4943 Arabic words and 119 other annotations correspond to Arabic numerals and isolated letters extracted from 26 pages of documents from the RASM2019 dataset, these pages extracted from two different books. The reference book "Add MS 23494", treatise on music theory relating to musical modes, prophets, signs of the zodiac, hours of the day, etc. and the reference book "Add MS 7474" treatise on the most influential and enduring astronomical subject to survive in the ancient world. The following table shows an overview of the number of pages extracted and other related information.

\vspace{0.5cm}

\begin{table} [h]
  \centering
\begin{tabular}  { | c | c| c |c| }
\hline
  & Book1 & BOOK 2 & Total  \\
\hline
Book reference        & "Add MS 23494"  &  "Add MS 7474" & 2  \\
\hline
Quantity of pages & 13  &  13 & 26  \\
\hline
Quantity of main text lines    & 257  &  240 & 497 \\
\hline
Quantity of words &  2472  &  2471 &  4943  \\
\hline
Quantity of isolated letter  &  9  &  87 &  96 \\
\hline
Quantity of Arabic numerals &  0  &  23 &  23 \\
\hline
 
\end{tabular}
    \caption{Some statistics on KALIMA-Dataset. }
    \label{tab:tab1}
\end{table}

Many of the text line images contain Arabic numerals, isolated letters belonging to geometric notations in documents, dealing with mathematical or astronomical subjects, as well as the presence of additional symbols and explanatory or commentary words outside the main line text that are positioned above or below the main words. In our dataset, at the line level, we ignore all words outside the main text line and annotate only the words in the main text, we do not annotate the comment words because they are located outside the main text line, which most of them are likely to be cut off when the line segmentation is performed. \textbf{ Figure 2 } Illustrates certain text line from ancient manuscript with different symbols, number and geometric letter which are not considered as words, as well as peripheral words out of the main text line. 

We put the annotation in two csv files, the first one contains the coordinates of each word in relation to the text line and the manuscript page, as well as the ground truth transcription and the order of each word in the corresponding line. A second csv file is provided including the position of each line in relation to the document page. Our dataset is availaible in the fllowing link \normalsize { \url{   https://github.com/bouchalhakim/KALIMA_Dataset.git}  }

\begin{figure}[H]
    \centering
    \includegraphics [width=0.9\linewidth] {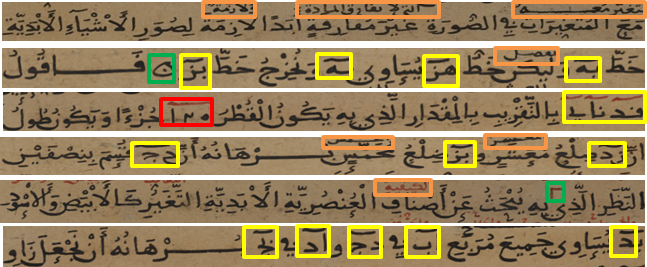}
    \caption{illustrate different forms of writing that do not belong to the main text line or are not considered words. The red, yellow, green and orange boxes contain Arabic numerals, geometric notations, symbols and additional words respectively.}
    \label{fig:2}
\end{figure}

\vspace{0.5cm}
We project all the set of grayscale word images of size 64x64 on a 2-dimensional plane, using the dimensionality reduction technique t-sne \cite{tsne} calculating the similarity between the images, we visualize well the placement of the images in two groups show that these images come from two distinct sets.

\begin{figure}[H]
    \centering
    \includegraphics [width=0.7\linewidth,height=0.7\linewidth] {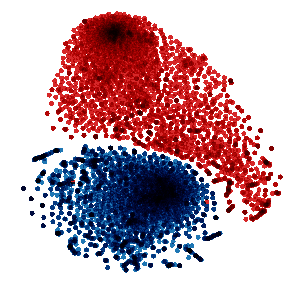}
    \caption{result of the projection of words on a 2 dimensional plane with the T-sne algorithm,the red dot corresponds to words in book1 and blue dot to book2.}
    \label{fig:3}
\end{figure}

\section{Conclusion}
\hspace{0.5 cm} In this paper, the KALIMA-Dataset is proposed  as a new dataset for handwritten Arabic words from ancient documents. It is the first established dataset in historical Arabic documents literature that annotates the word position with bounding boxes. This dataset can be used to develop and test systems of extraction and recognition of ancient Arabic words. Each word is accompanied by position coordinates, numerical transcription of words in Arabic, word position order in the line image. As perspective we plan to add more pages of old documents from different sources annotated at word level.


\begin{thebibliography}{100}
\bibitem{article123}
C. B. Rasmussen, K. Kirk, and T. B. Moeslund. The Challenge of Data Annotation in Deep Learning-A Case Study on Whole Plant Corn Silage.  Sensors, vol. 22, no. 4, p. 1596, Feb. 2022, doi: 10.3390/s22041596.

\bibitem{article2}
A. Al-Dmour and R. Zitar. Word Extraction from Arabic Handwritten Documents Based on Statistical Measures. 
Int. Rev. Comput. Softw., Jun. 2016.

\bibitem{article3}
N. Aouadi and A. Echi. Word Extraction and Recognition in Arabic Handwritten Text.
Int. J. Comput. Inf. Sci., vol. 12, pp. 17-23, Sep. 2016, doi: 10.21700/ijcis.2016.103.

\bibitem{article4}
J. H. AlKhateeb, J. Jiang, J. Ren, and S. S. Ipson. Component based Segmentation of Words from Handwritten Arabic Text. 
p. 5, 2009.

\bibitem{article5}
R. Ahmad, S. Naz, M. Z. Afzal, S. F. Rashid, M. Liwicki, and A. Dengel. KHATT: A Deep Learning Benchmark on Arabic Script. 
in 2017 14th IAPR International Conference on Document Analysis and Recognition (ICDAR), Kyoto, Nov. 2017, pp. 10-14. doi: 10.1109/ICDAR.2017.358.

\bibitem{article6}
C. Clausner, A. Antonacopoulos, N. Mcgregor, and D. Wilson-Nunn. ICFHR 2018 Competition on Recognition of Historical Arabic Scientific Manuscripts - RASM2018. in 2018 16th International Conference on Frontiers in Handwriting Recognition (ICFHR), Niagara Falls, NY, USA, Aug. 2018, pp. 471-476. doi: 10.1109/ICFHR-2018.2018.00088.

\bibitem{IBN SINA} 
Reza Farrahi Moghaddam, Mohamed Cheriet, Mathias M. Adankon, Kostyantyn Filonenko, and Robert Wisnovsky. IBN SINA: A database for research on processing and understanding of Arabic manuscripts images. Proceedings of DAS\textquotesingle 10, June 9-11, 2010, Boston, MA, USA 
 
\bibitem{VMLHD} 
M. Kassis, A. Abdalhaleem, A. Droby, R. Alaasam and J. El-Sana, VML-HD: The historical Arabic documents dataset for recognition systems.  2017 1st International Workshop on Arabic Script Analysis and Recognition (ASAR). 2017, pp. 11-14, doi: 10.1109/ASAR.2017.8067751.

\bibitem{MHDID} 
A. Shahkolaei, A. Beghdadi, S. Al-maadeed and M. Cheriet.  MHDID: A Multi-distortion Historical Document Image Database.  2018 IEEE 2nd International Workshop on Arabic and Derived Script Analysis and Recognition (ASAR), 2018, pp. 156-160, doi: 10.1109/ASAR.2018.8480372

\bibitem{BADAM} 
Benjamin Kiessling, Daniel St{\"o}kl Ben Ezra, Matthew Thomas Miller. BADAM: A Public Dataset for Baseline Detection in Arabic-script Manuscripts. In: Proceedings of the 5th International Workshop on Historical Document Imaging and Processing (HIP). pp. 13-18 (2019)

\bibitem{tsne} 
 Melit Devassy B, George S. Dimensionality reduction and visualisation of
hyperspectral ink data using t-SNE. Forensic Sci Int (2020) 311:110194.
doi: 10.1016/j.forsciint.2020.110194

\end{thebibliography}
\end{document}